%% file: main.tex
\documentclass[sigconf]{acmart}

\def\BibTeX{{\rm B\kern-.05em{\sc i\kern-.025em b}\kern-.08emT\kern-.1667em\lower.7ex\hbox{E}\kern-.125emX}}

\usepackage{amsmath,bm}
\usepackage{bbm}
\usepackage{booktabs}
\usepackage{float}

\DeclareMathOperator*{\argmax}{argmax}
\DeclareMathOperator*{\argmin}{argmin}
\DeclareMathOperator*{\mean}{mean}

\acmConference[Submitted to KDD '19]{Submitted to KDD '19}{August 04--08, 2019}{Anchorage, AK}
\begin{document}

%
% The "title" command has an optional parameter, allowing the author to define a "short title" to be used in page headers.
\title{Dynamic Pricing for Airline Ancillaries with Customer Context}

\author{Naman Shukla}
\affiliation{%
  \institution{University of Illinois at Urbana-Champaign}
  \city{Urbana}
  \state{IL}}
\email{namans2@illinois.edu}

\author{Arinbj\"orn Kolbeinsson}
\affiliation{%
  \institution{Imperial College London}
  \city{London}
  \country{UK}}
\email{ak711@imperial.ac.uk}

\author{Ken Otwell}
\affiliation{%
  \institution{Otwell Research}
  \city{Huntsville}
  \state{AL}}
  \email{ken.otwell@gmail.com}

\author{Lavanya Marla}
\affiliation{%
  \institution{University of Illinois at Urbana-Champaign}
  \city{Urbana}
  \state{IL}}
\email{lavanyam@illinois.edu}
  
\author{Kartik Yellepeddi}
\affiliation{%
  \institution{Deepair solutions}
  \city{London}
  \country{UK}}
\email{kartik@deepair.io}

\renewcommand{\shortauthors}{Shukla et al.}

%
% The abstract is a short summary of the work to be presented in the article.
\begin{abstract}
    Ancillaries have become a major source of revenue and profitability in the travel industry. Yet, conventional pricing strategies are based on business rules that are poorly optimized and do not respond to changing market conditions.
    
    This paper describes the dynamic pricing model developed by Deepair solutions, an AI technology provider for travel suppliers. We present a pricing model that provides dynamic pricing recommendations specific to each customer interaction and optimizes expected revenue per customer. The unique nature of personalized pricing provides the opportunity to search over the market space to find the optimal price-point of each ancillary for each customer, without violating customer privacy.
   
    In this paper, we present and compare three approaches for dynamic pricing of ancillaries, with increasing levels of sophistication: (1) a two-stage forecasting and optimization model using a logistic mapping function; (2) a two-stage model that uses a deep neural network for forecasting, coupled with a revenue maximization technique using discrete exhaustive search; (3) a single-stage end-to-end deep neural network that recommends the optimal price. We describe the performance of these models based on both offline and online evaluations. We also measure the real-world business impact of these approaches by deploying them in an A/B test on an airline's internet booking website. We show that traditional machine learning techniques outperform human rule-based approaches in an online setting by improving conversion by 36\% and revenue per offer by 10\%. We also provide results for our offline experiments which show that deep learning algorithms outperform traditional machine learning techniques for this problem. Our end-to-end deep learning model is currently being deployed by the airline in their booking system.
    
    % \vfill\null
    % \vspace*{80px}
    % \columnbreak
\end{abstract}



%
% This command processes the author and affiliation and title information and builds
% the first part of the formatted document.
\maketitle

%%%%%%%%% BODY TEXT
\input{sections/1_introduction.tex}
\input{sections/2_pricingfactors.tex}

\input{sections/3_pricingsystem.tex}

\input{sections/4_priceeval.tex}
\input{sections/5_Training}
\input{sections/6_evaluation.tex}
\input{sections/7_discussion.tex}
\input{sections/8_conclusion.tex}

\begin{acks}
We sincerely and gratefully acknowledge our airline partners for their continuing support.
\end{acks}

%
% The next two lines define the bibliography style to be used, and the bibliography file.
\bibliographystyle{ACM-Reference-Format}
\bibliography{main}

\end{document}

%% file: sections/1_introduction.tex
\section{Introduction} \label{Introduction}

Ancillaries are optional products or services sold by businesses to complement their primary product \cite{bockelie2017incorporating}. In the airline industry, these services or products can be directly related to a passenger's flight itinerary, such as baggage allowance, leg room, seat upgrades or meals, or may be related to the passenger's overall travel plan, for example, hotel rooms, rental cars, or destination activities. The estimated ancillary revenue collected by major US air carriers was more than \$18 billion in 2015, and \$59 billion for airlines around the world in the same year \cite{IdeaWorks}. 

Even though this revenue stream is clearly substantial to the airline industry, its pricing strategies are not fully developed due to its recent emergence in the market. Because these products were traditionally not offered as ancillaries, airlines have little knowledge of the relationship between the customers' choice of primary product and the ancillary product. Moreover, given the now optional nature of these products, ancillary purchases are a result of deep personal preferences of each individual and the context of their trip. Consequently, airlines experience very low conversion rates (less than 5\%) for ancillaries. Understanding these personal preferences based on the context of each shopping session is crucial to pricing them effectively and generating revenue. Moreover, various ancillaries compete with each other's "shelf-space" on the website and wallet-share of the customer, so pricing an ancillary in the context of other ancillaries confounds the problem.

Currently, the majority of ancillary products are static price-points, i.e., invariant to customer or itinerary characteristics. Our aim is to develop a price recommendation system specific to ancillary services, to price these products dynamically based on itinerary-specific information.

\begin{figure}[ht]
\centering
\includegraphics[scale=0.7]{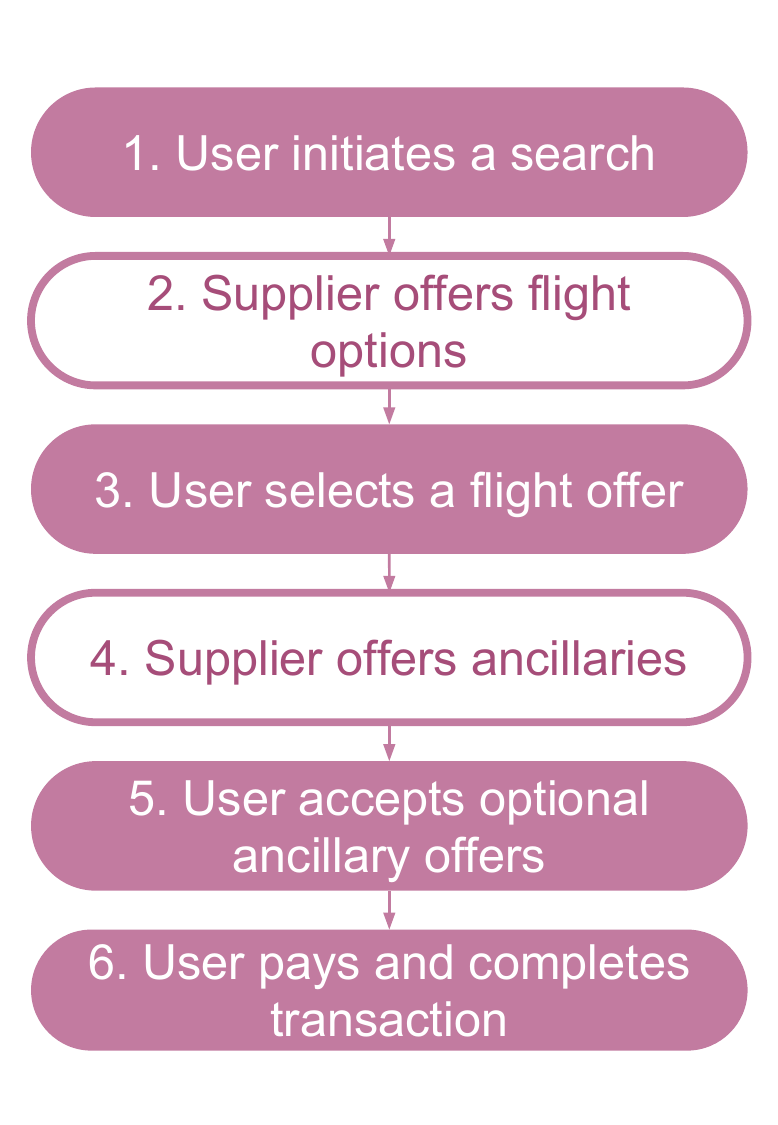}
\caption{User live session as state space}
\label{fig:ss}
\end{figure}

Each \emph{booking session} on an airline's website can be modeled using the state space represented in Figure \ref{fig:ss}. The first three steps involve the primary product, which is a set of seats on aircraft connecting the origin to the destination, referred to as \emph{right-to-fly}; while ancillary offerings and corresponding customer choice occur from state $4$ onwards. Note that in this work we consider only ancillaries offered during the booking session, and not those booked later, such as adding bags after reaching the airport. The goal of this work is to optimize prices dynamically, while estimating willingness to pay; and therefore we will address the latter case in future work.

\begin{figure}[h]
\centering
\includegraphics[scale=0.48]{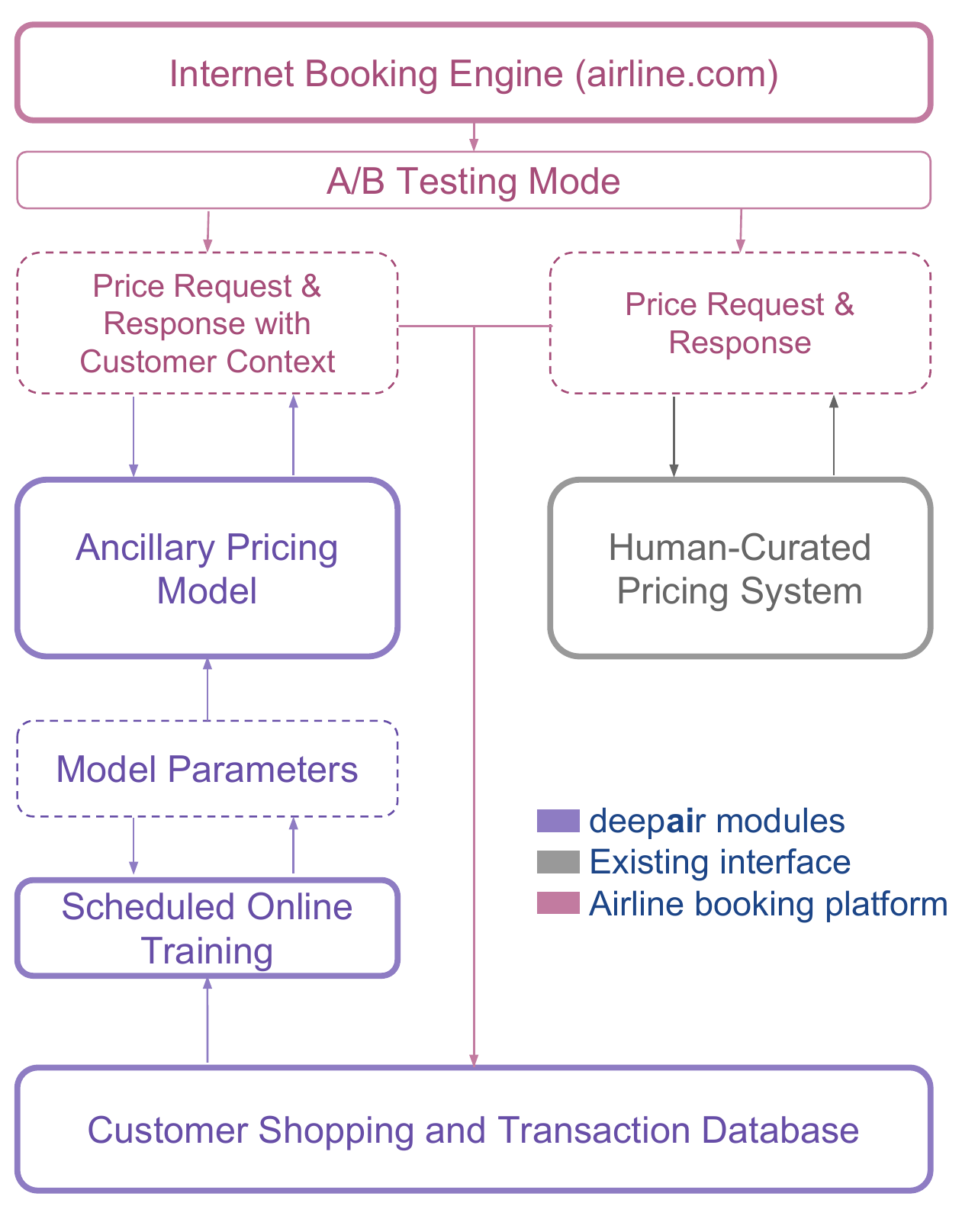}
\caption{System integration with existing pipeline}
\label{fig:si}
\end{figure}

Conventional pricing frameworks are static, and not capable of recommending a price conditioned upon rich session-specific information. Our pricing suggestions are generated through an A/B testing framework that directs live booking traffic to various deployed models. The integration specifications for inference and data retrieval for training, with respect to the current pipeline at an airline booking engine, are described in Figure \ref{fig:si}.

Our contributions are as follows.
\begin{itemize}
\item We present a rich, customized, session-specific dynamic price recommendation system for ancillary services that significantly outperforms existing pricing systems in terms of revenue.
\item We develop a deep learning model that effectively estimates purchase probability and \emph{simultaneously} prices the ancillary product, by modeling monotonicity properties of customers' willingness to pay. This model provides improved revenues and captures customers' behavior more accurately than sequential models that combine traditional machine learning (or deep learning) models followed by revenue optimization.
\item Our model predicts human choice more accurately than our baseline model, resulting in increased conversion for the ancillary product.
\item We implement and test our models on real data, both on historical data and by live testing in an airline's booking system, and demonstrate both offline and online improvement based on live customer usage statistics.
\end{itemize}

\subsection{Related Work}
Unbundling is the process of separating a product into primary and ancillary products to allow customers more flexibility of purchase, and businesses to increase revenues by matching customer needs more accurately. In the airline industry, this phenomenon has been led by low-cost carriers (LCCs), whose operational and pricing models rely heavily on ancillary fees. In recent years, many legacy airlines have adopted this strategy \cite{Garrow}. Despite an initial response through negative emotions and retaliatory behavior \cite{FreeToFee}, unbundling of services into the basic right-to-fly and additionally priced ancillary services (bags, meals, etc.) has gradually gained acceptance among customers \cite{AcceptanceAncillary}. In fact, revenue from ancillary services in the airline industry have nearly tripled in the past decade, from 3\% to 8\% of total revenue \cite{Ancillary_OliverWyman}.

Studies on this new phenomena in the airline context are ongoing \cite{CuiEtAl,AllonEtAl}. Economics literature indicates that the practice of offering add-ons (an equivalent term for ancillaries) can raise equilibrium profits when airlines compete; and can also be used for price discrimination and customer segmentation \cite{Ellison}. Allon et al\cite{AllonEtAl} argue that unbundling and baggage fees are consistent with reduction of airline operating costs, but may not effectively segment customers. Customer characteristics and the airline's ability to price discriminate are also shown to significantly influence its profits \cite{CuiEtAl}. Bockelie and Belobaba \cite{bockelie2017incorporating} study behavioral models of ancillary product purchase, and specifically comparing the difference in price perceptions of customers who purchase ancillary services sequentially or simultaneously. While multiple behavioral theoretical models \cite{KahnemanTversky,GabaixLaibson,ShulmanGeng} based on risk perception, knowledge levels and bounded rationality; and discrete choice models \cite{BenAkiva} are typically used to model customer choice, there is limited literature that explicitly models the relationship between ancillary services and the primary product (itinerary, or fare class).   

Topics in dynamic pricing of homogeneous products have been extensively studied \cite{den2015dynamic}. In fact, dynamic pricing has been a catalyst for innovation in various transport and service industries. Ride-hailing platforms have used surge pricing to match demand and supply, and to avoid the "wild-goose chase" problem \cite{castillo2017surge}. Related to our problem is the work of \citet{ye2018customized} for Airbnb accommodation pricing. They formulate a custom scheme to optimally price each product using a triple-stage model with booking probability classification, price-suggestion regression and seller-specific logic. Whereas Airbnb considers all their listings as unique and all the customers identical, we consider the inverse problem of identical products and unique customers.

%% file: sections/2_pricingfactors.tex
\section{Pricing Factors} \label{Pricing Factors}
In this section, we discuss the two primary factors we use to determine the optimal price for an ancillary: the demand function and customer attributes.

\subsection{Demand Function} \label{Demand Function}
An estimation of a demand curve $D(P)$ as a function of price $P$, can be obtained by evaluating the variation of demand with respect to price. Then, the optimal price can be obtained via maximizing the expected revenue based on the estimated demand curve. The optimal value can only be obtained when the demand function $D(P)$ is an accurate estimate of the actual demand in the market, else the $P^*$ and corresponding revenue will be sub-optimal.

\begin{equation}
P^* = \argmax_P P \times D(P)\label{eq:demand}
\end{equation}

For airline ancillaries, the demand function $D$ is not just a function of price $P$ offered but also of customer attributes $\mathbbm{x}$. Hence, a better estimation of the demand function is $D(P, \mathbbm{x})$ and can be obtained by observing the change in demand conditioned on both price $P$ and customer attributes $\mathbbm{x}$. In our study, we implement and compare two algorithms to estimate the probability of a customer purchasing an offered ancillary, which we assume to be a proxy for estimated demand $D(P, \mathbbm{x})$. Details of our algorithms are presented in Section \ref{Pricing Models}.

\subsection{Customer Attributes} \label{Customer Attributes}

We define a customer's attributes, $\mathbbm{x}$, as the set of factors that influences the probability of that customer purchasing the offered ancillary, \emph{at a given price}. The major attributes that the demand function is found to be significantly dependent on are time, market, items already in the cart and length of stay.

\subsubsection*{Time}
There are two types of time-related factors that heavily influence the demand function:
(1) Days to departure: Price sensitivity captures the relationship between the price and the propensity for purchasing the product. Usually, customers who buy their tickets far in advance are more price sensitive than customers who buy closer to departure.
(2) Departure date and time: Like the right-to-fly, ancillary demand has strong time-of-day and seasonal variations. Itineraries starting on certain days and times have higher `quality' and hence increased demand; due to factors such as higher convenience of time of travel, better connectivity (neither too long nor too short connection time), better availability of alternative connections, special events, and holidays. The quality of service has a strong correlation with the type of passengers it attracts. It is well-known that low quality services tend to be cheaper and attract more price sensitive customers.

\begin{figure}[h]
\centering
\includegraphics[scale=0.18]{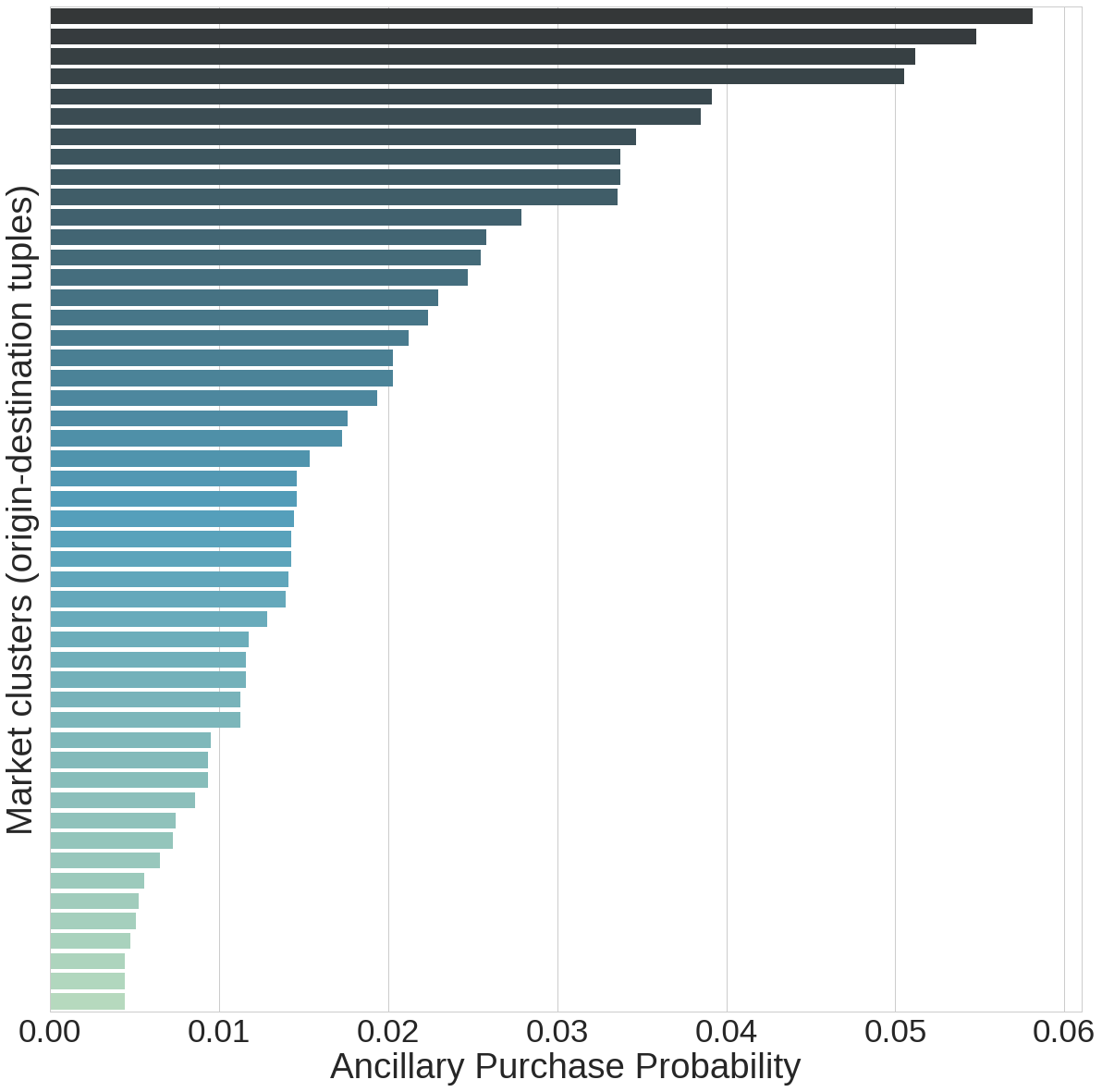}
\caption{Market clusters' probability of ancillary purchase}
\label{fig:routes}
\end{figure}

\subsubsection*{Markets}
Airlines serve a large variety of markets. A market is a tuple comprising of the origin and destination of the trip. Certain markets are served with a larger fraction of non-stop itineraries, while others may be served with larger fractions of itineraries containing connections. Certain markets have a heavier fraction of business trips, while others consist primarily of leisure trips. As shown in Figure \ref{fig:routes}, there are clusters of markets that have high demand for ancillary services as compared to others. To estimate the demand using \eqref{eq:demand}, we segment these clusters into \emph{sub-markets}. We define \emph{sub-markets} as a mapping from a vector of customer attributes $\mathbbm{x}$ to an origin-destination cluster where estimated demand is statistically similar for a given prior ancillary price. 

\subsubsection*{Length of Stay}
\begin{figure}[h]
\centering
\includegraphics[scale=0.18]{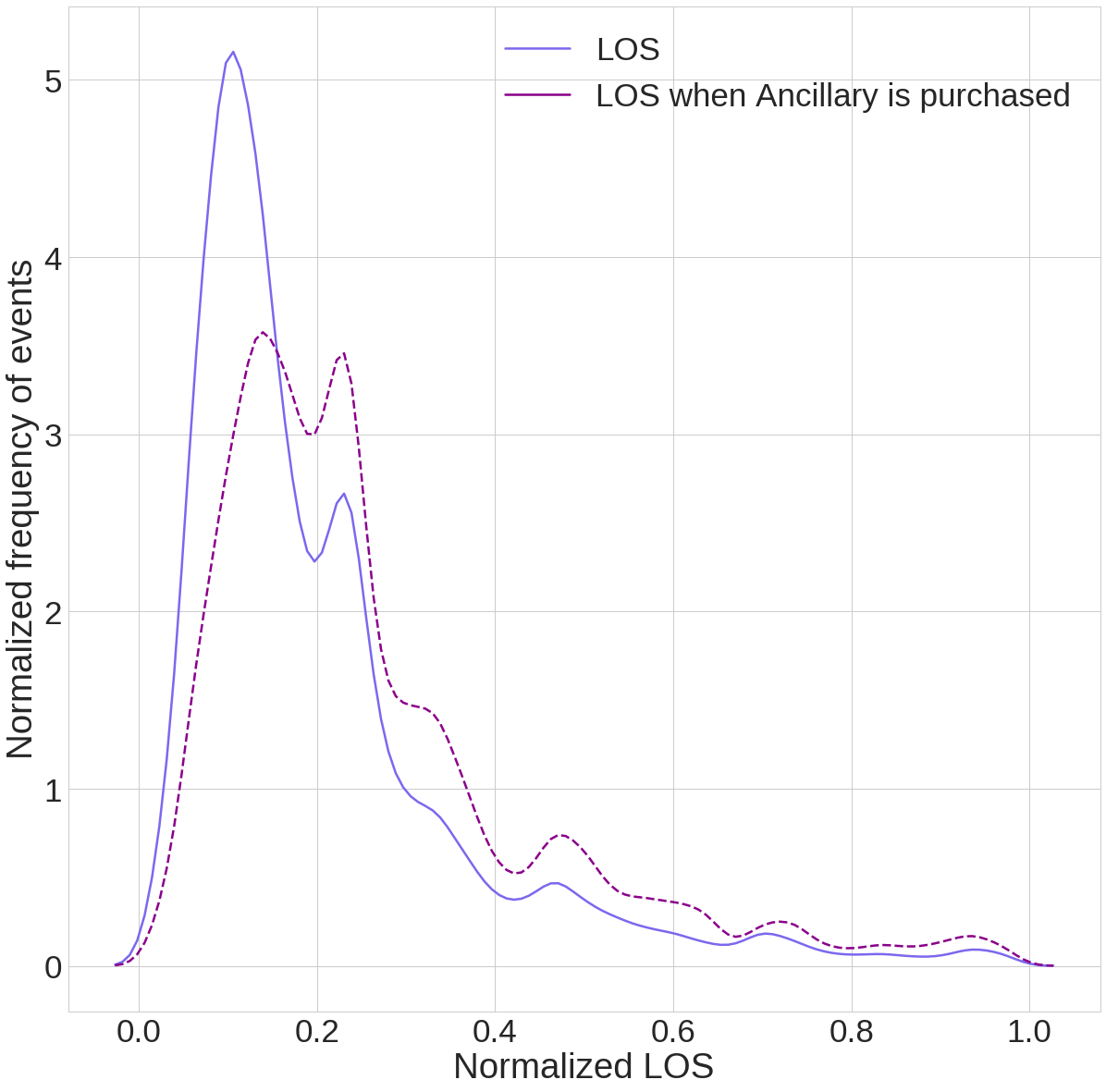}
\caption{LOS signal from KDE}
\label{fig:los}
\end{figure}

For those passenger bookings that are round-trips, we define Length of stay (LOS) as the number of days a passenger plans to stay at the destination. If the passenger does not have a return ticket we consider LOS to be 0. Figure \ref{fig:los} shows estimated kernel density function for LOS for two types of bookings: when an ancillary was purchased (dashed line), and for all bookings (solid line). These estimated graphs are irregular from normalized LOS values of 0.0 to 0.3 (approximately), indicating higher chances of ancillary purchase for LOS durations that are neither too short nor too long. These irregularities indicate that passengers prefer to purchase ancillaries (such as bags), for medium length trips for which they might require additional storage space. Hence, the apparent signature describes the conditional importance of the LOS attribute.

%% file: sections/3_pricingsystem.tex
\section{Pricing Models} \label{Pricing Models}
Our pricing model consists of two components: (i) an ancillary purchase probability model that is structured as a binary classification problem, and (ii) a revenue optimization model that, given the probability of purchase, recommends an optimal price that maximizes the airline's expected revenue. 

We make the following assumptions:

\begin{itemize}
\item \textit{Pricing range}: The recommended ancillary price is allowed to vary only within a legal range defined by business strategy as mentioned in section \ref{Experiments}. 
% \lmcomment{user here is the airline, not the passenger, correct?}
\item \textit{Monotonicity in willingness to pay}: If a customer is willing to purchase a product for price $p$, they are willing to purchase the same product at a price $p' < p$. Similarly, if a customer is unwilling to purchase a product at price $p$, they will be unwilling to purchase at a price $p' > p$. 
%\item \textit{Convexity}: The expected revenue function is convex in nature and has a unique maximum.
\end{itemize} 

\begin{figure}[h]
\centering
\includegraphics[width=0.4\textwidth]{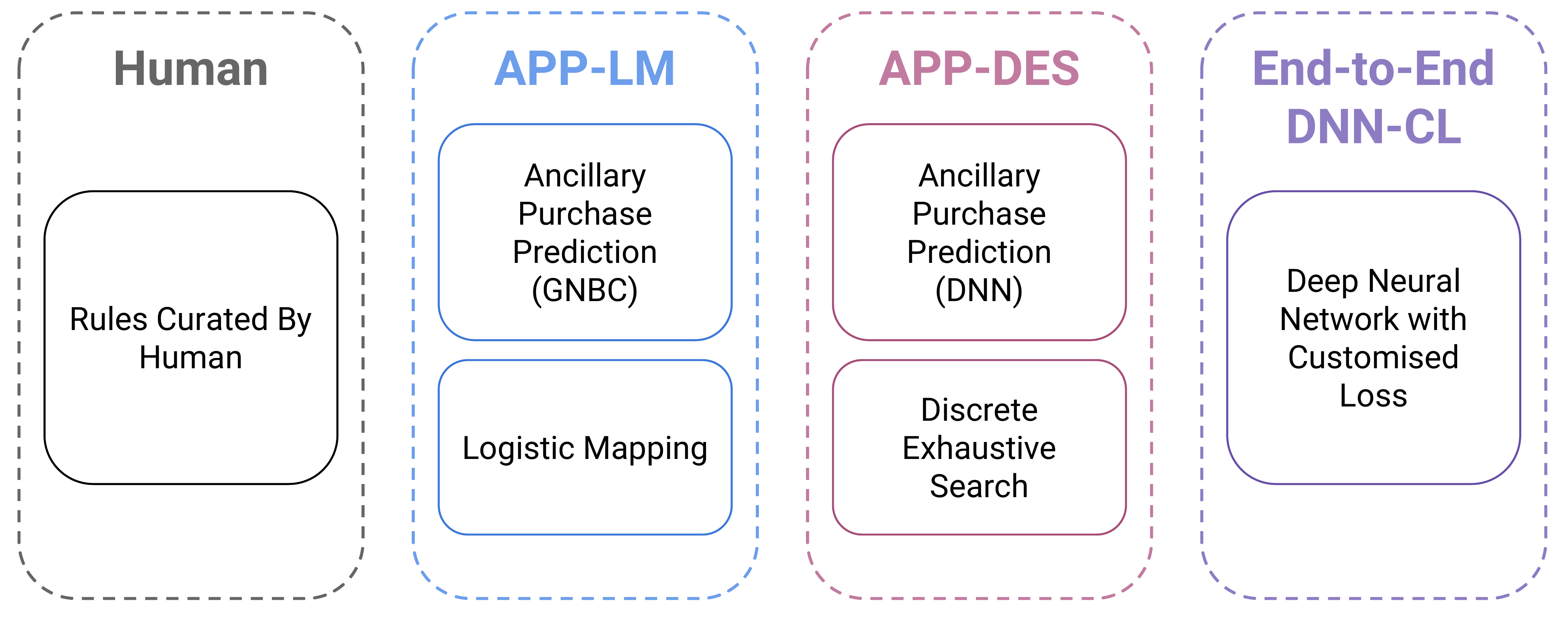}
\caption{Schematic of Ancillary Pricing Models}
\label{fig:overview}
\end{figure}

We implement three different pricing models, of increasing complexity, shown in Figure \ref{fig:overview}. These are embedded into the framework in Figure \ref{fig:si}. 
\begin{enumerate}
\item \textbf{Ancillary purchase prediction with logistic mapping (APP-LM)}: This model uses a Gaussian Naive Bayes with clustered features (GNBC) model for ancillary purchase probability prediction and a pre-calibrated logistic price mapping function for revenue optimization. 
\item \textbf{Ancillary purchase prediction with exhaustive search (APP-DES)}: This model uses a Deep-Neural Network (DNN) trained using a weighted cross-entropy loss function for ancillary purchase probability estimation. For price optimization, we implement a simple discrete exhaustive search algorithm that finds the optimal price point within the pricing range. 
\item \textbf{End-to-End DNN with custom loss function (DNN-CL)}: This DNN-based model is trained on a customized loss function, and presented in section \ref{Customized Loss Function}. This loss function is designed using the strategic model objective function\cite{ye2018customized} and explicitly models the dominance properties embedded within the willingness to pay assumption. 
\end{enumerate}
In APP-LM and APP-DES, the ancillary purchase probability model and the revenue optimization model are sequential whereas in DNN-CL, they are simultaneously solved to achieve the recommended price.

\subsection{Ancillary Purchase Probability Model} \label{Ancillary Purchase Prediction Model}
Our ancillary purchase probability model estimates the demand curve within a sub-market for each offered ancillary, for a given price of the ancillary. This is formulated as a binary classification problem. We aim to estimate the probability distribution function $f_\theta(\bm{x}, P)$, where $\bm{x}$ $|$ $\bm{x}\subseteq\mathbbm{x}$,  is the feature vector and $P$ is the offered price. We used over 30 features that fall under the following categories:

\begin{itemize}
\item Temporal features : Length of stay, seasonality (time of the day, month of the year, etc), time of departure, time of shop, time to departure.
\item Market-specific features : Arrival and destination airport, arrival and destination city, ancillary popularity for the route, etc.
\item Price comparison scores : Scores based on alternative/same flights across/within the booking class.
\item Journey specific features : Group size, booking class, fare group, number of stops, etc.
\end{itemize}

As mentioned earlier, the binary classification task is highly challenging because ancillary purchase is highly imbalanced (class ratios of $6:100$). For APP-LM, We first experiment with many traditional classification algorithms like Gaussian Naive Bayes (GNB), Gaussian Naive Bayes with clustered features (GNBC), Random Forest (RF), using features chosen based on principal component analysis for these algorithms \cite{scholkopf1998nonlinear}.

For APP-DES, we use a customized deep neural network (DNN) trained on weighted cross entropy loss, as a classifier. While the DNN did not require a lot of feature engineering, we  experimented with various hyper-parameters like network architecture, drop-out rates, activation functions, optimization algorithms, and convergence criteria.
\pagebreak
\subsection{Revenue Optimization} \label{Revenue Optimization}

\subsubsection*{Logistic Price Mapping Function} \label{Logistic Price Mapping Function}
In our base model APP-LM, once the ancillary purchase probability is predicted, we use a logistic function to recommend a price. The intuition behind using logistic mapping is that the ancillary can be priced closer to the maximum of the pricing range when the probability of purchase is high, and lower for low probabilities. Hence, a price mapping is chosen based on \eqref{eq:logistic}.

\begin{equation}
P^{rec} = \frac{L}{1+ \exp{^-k(x-x_0)}}
\label{eq:logistic}
\end{equation}

\begin{figure}[h]
\centering
\includegraphics[width=0.45\textwidth]{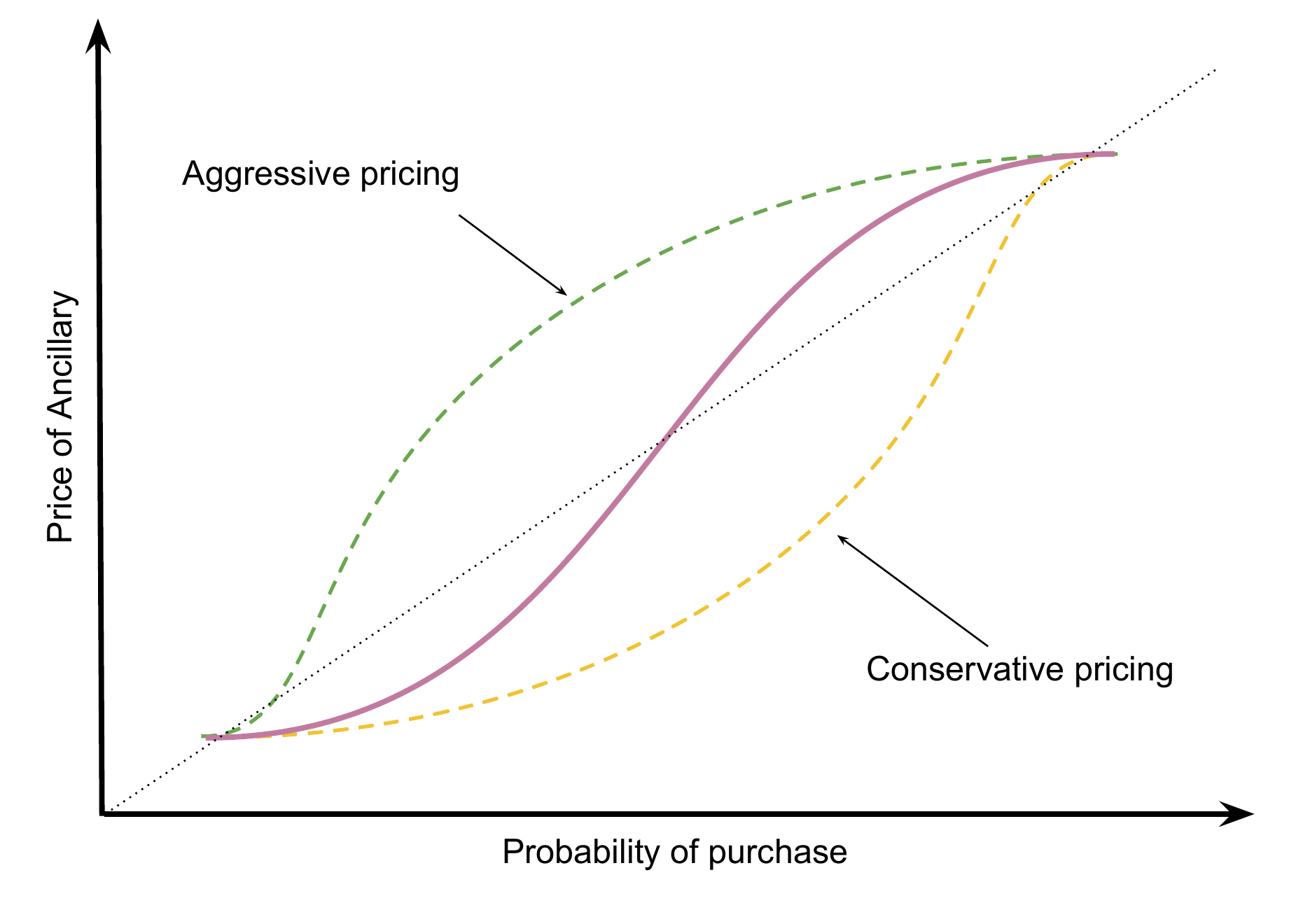}
\caption{Logistic mapping from probability of purchase to a recommended price.}
\label{fig:logistic}
\end{figure}

According to \eqref{eq:logistic}, three parameters can be controlled to map the price desirably. 

\begin{itemize}
\item Max value, $L$ : this is the full price of the ancillary
\item Shape factor, $k$ : the shape or steepness of the curve
\item Mid point, $x_0$ : the mid point of the sigmoid curve
\end{itemize}

The shape factor $k$ and mid-point $x_0$ can be fine-tuned to be either aggressive or conservative with pricing. This tuning is illustrated in Figure \ref{fig:logistic}, indicating that at low purchase probabilities, the model compensates by reducing the  recommended price.

\subsubsection*{Discrete Exhaustive Search} \label{Discrete Exhaustive Search}

Exhaustive search can be efficiently performed over a small set of discrete prices that are within the pricing range. For a given probability of purchase $f_{\theta}(\bm{x}, P)$ and price $P$, expected revenue is computed using \eqref{eq:exp_rev}. 

\begin{figure}[h]
\centering
\includegraphics[width=0.45\textwidth]{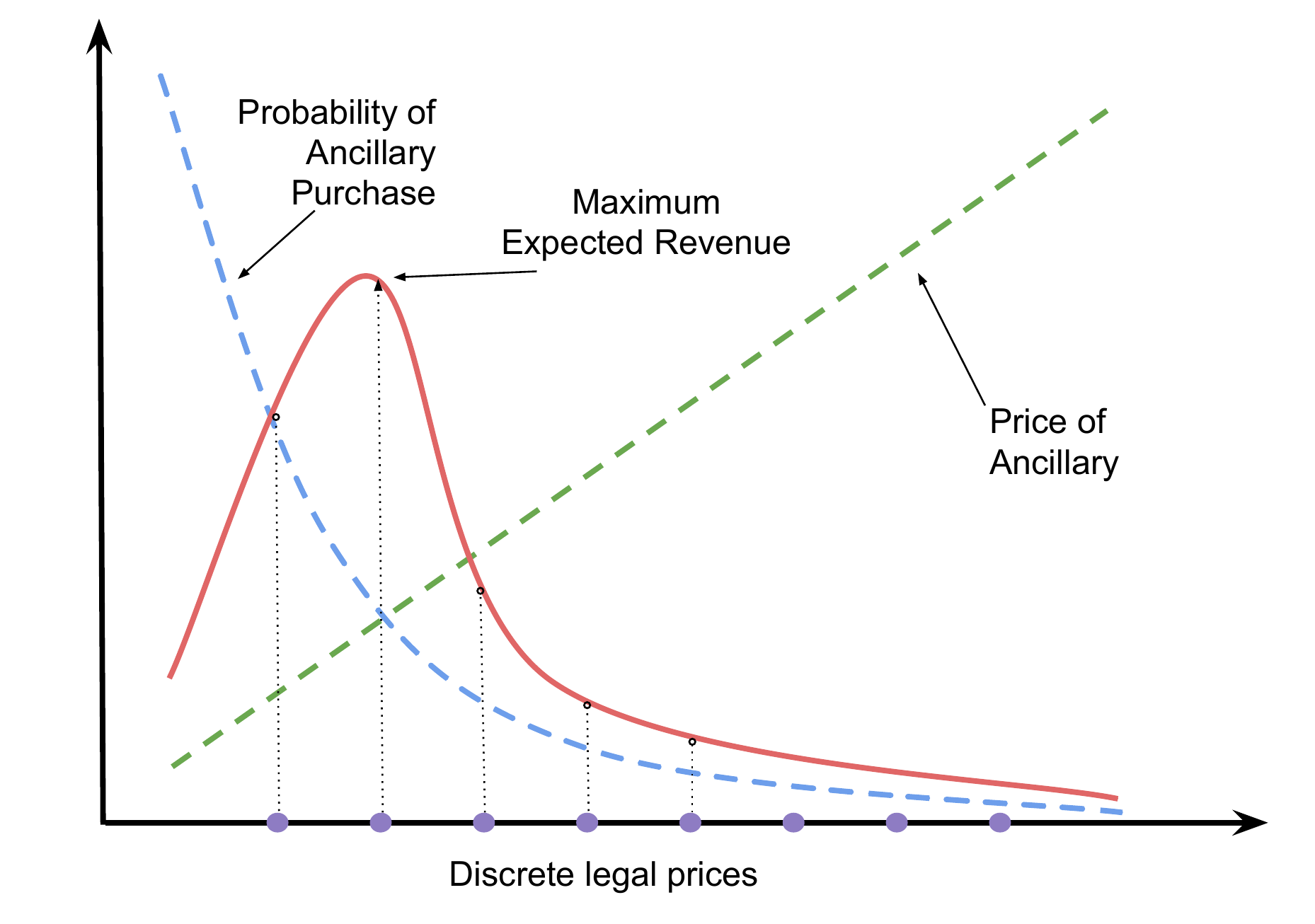}
\caption{An illustration of a discrete search in the price range}
\label{fig:des}
\end{figure}

\begin{equation}
\hat{\mathbbm{E}}_P = P \times f_{\theta}(\bm{x}, P)
\label{eq:exp_rev}
\end{equation}

Without assuming that the revenue function is convex but only unimodal, an exhaustive search over all allowed prices can be performed, as represented in figure \ref{fig:des}. The convexity assumption is met only when price sensitivity has a small derivative in the region of interest. Thereby, using exhaustive search, the optimal price can be evaluated using equation \ref{eq:des_eq}. As discussed in Section \ref{Demand Function}, the optimality of the price $P^{rec}$ is dependent on the accuracy of estimation of the demand.

\begin{equation}
P^{rec} = \argmax_P \hat{\mathbbm{E}}_P
\label{eq:des_eq}
\end{equation}

The performance of a two-stage sequential forecasting and optimization method (like APP-LM and APP-DES), depends on a good demand estimate over the permissible range of prices. This requires sufficient exposure of those prices in the market to learn an accurate price sensitivity curve for each sub-market. Without such data, approximate methods such as custom loss functions can produce more revenue in practice.

\subsection{Customized Loss Function for DNN-CL} \label{Customized Loss Function}
In this section, we present a customized loss function that takes into account a regret of pricing low, conditional on the ancillary being purchased; and a penalty for recommending high, conditional on it not being purchased. The objective function is inspired from the strategic model proposed by \citet{ye2018customized} and $\epsilon$-insensitive loss used in SVR\cite{smola2004tutorial}. We enhance this strategic model using latent variables to incorporate the monotonicity in the willingness to pay assumption in our loss function. Suppose we are given $N$ training samples $\{\bm{x}_i, y_i\}_{i=1}^{N}$, where $\bm{x}_i$ is the feature vector and $y_i$ is the ground truth label for the $i^{th}$ session. For purchased ancillaries, $y_i$ equals 1 and 0 otherwise. The recommended price $P^{rec}$ for feature vector $\bm{x}$ is denoted by $P^{rec} = \mathbbm{F}_\Theta(\bm{x}, \mathbbm{P})$, where $\Theta$ is a set of trainable parameters that can be learned for the mapping function $\mathbbm{F}$, and $\mathbbm{P}$ is a set of discrete price points in the pricing range.

\begin{figure}[]
\centering
\includegraphics[scale=0.5]{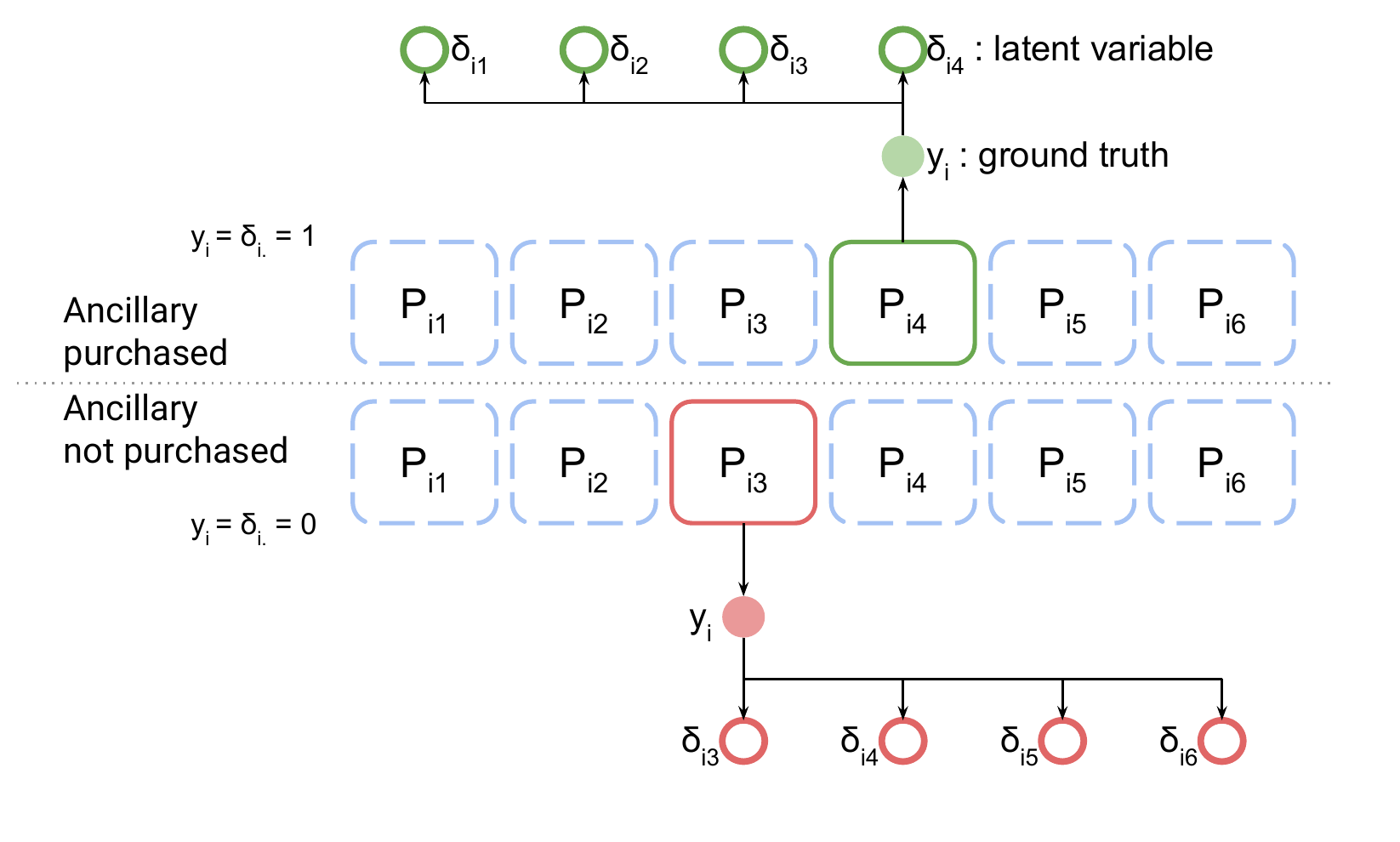}
\caption{Latent variable $\delta$ mapping from ground truth $y$}
\label{fig:delta}
\end{figure}

The objective of the learning is to minimize the loss $\bm{\mathcal{L}}$ given as

\begin{equation}
\bm{\mathcal{L}} = \argmin_{\theta} \sum_{i=1}^{N} \sum_{j=1}^{|\mathbbm{P}|} (\Phi_{lb} + \Phi_{ub})\cdot \mathbbm{1}_{(\sigma_{ij} > 0)}
\end{equation}

where the lower bound function $\Phi_{lb}$ and upper bound function $\Phi_{ub}$ are defined as, 

\begin{align*}\label{eq:pareto mle2}
\Phi_{lb} = \max \bigg(0, \Big( L(P_{ij}, \delta_{ij})-\mathbbm{F}_\Theta(\bm{x}_i, \mathbbm{P})\Big) \bigg) \\
\Phi_{ub} = \max \bigg(0, \Big( \mathbbm{F}_\Theta(\bm{x}_i, \mathbbm{P}) - U(P_{ij}, \delta_{ij})\Big) \bigg)
\end{align*}

where $\delta_{ij}$, shown in Figure \ref{fig:delta}, is a latent variable that ensures the monotonicity in the willingness to pay assumption by taking the current ground truth $y_{i}$ into account. The indicator function $\mathbbm{1}_{(\sigma_{ij} > 0)}$ selects loss values corresponding to those $\delta_{ij}$ which satisfy the monotonicity condition. Therefore, $\delta_{ij}$ is defined as   

\begin{equation}
\delta_{ij}(y_{i}) = \begin{cases}
y_{i} & \text{if $\sigma_{ij}\geq0$}\\
0 &\text{otherwise}
\end{cases}
\end{equation}

Where, $\sigma$ is the willingness to pay factor, defined as:

\begin{equation}
    \sigma_{ij} = (j - j^*)\cdot(-1)^{y_{i}}
\end{equation}

Assuming prices are listed in ascending order, $j^*$ is the index at which $P_{ij}$ equals $\mathbbm{F}_\Theta(\bm{x}_i, \mathbbm{P})$. We use $L$ and $U$ for the lower bound and the upper bound of the optimal price range, respectively. The functions $L(P_{ij}, \delta_{ij})$ and $U(P_{ij}, \delta_{ij})$ are defined as follows:

\begin{equation}
L(P_{ij}, \delta_{ij}) = \delta_{ij}\cdot P_{ij} + (1-\delta_{ij})\cdot c_1P_{ij}
\end{equation}

When the ancillary is purchased, the lower bound $L$ is the purchase price $P_{ij}$. Otherwise, a lower price of $c_1P_{ij}$ is set to be the lower bound, where $c_1 \in (0,1)$.  

\begin{equation}
U(P_{ij}, \delta_{ij}) = (1-\delta_{ij})\cdot P_{ij} + \delta_{ij}\cdot c_2P_{ij}
\end{equation}

The upper bound $U$ is $P_{ij}$ when the ancillary is not purchased,  whereas if the ancillary is purchased, a price of $c_2P_{ij}$ ($c_2 > 1$) is set as the upper bound. 

\begin{table}[h]
    \caption{Lower bound and Upper bound loss values}
    \begin{tabular}{@{}ccc@{}}
    \toprule
    Prices & $\Phi_{lb}\cdot \mathbbm{1}_{(\sigma_{ij} > 0)}$ & $\Phi_{ub}\cdot \mathbbm{1}_{(\sigma_{ij} > 0)}$ \\ \midrule
    $P_{ij} < \mathbbm{F}_\Theta$ & 0 & $\max(0, \mathbbm{F}_\Theta(\bm{x}_i, \mathbbm{P}) - c_2P_{ij})$ \\
    $P_{ij} = \mathbbm{F}_\Theta$ & 0 & 0 \\
    $P_{ij} > \mathbbm{F}_\Theta$ & $\max(0, c_1P_{ij}-\mathbbm{F}_\Theta(\bm{x}_i, \mathbbm{P}))$ & 0 \\ \bottomrule
    \end{tabular}
\label{tab:2}
\end{table}

Table \ref{tab:2} illustrates the lower and upper bound loss values for recommended price with respect to discrete price points. For $P_{ij} < \mathbbm{F}_\Theta(\bm{x}_i,\mathbbm{P})$, the upper bound loss increases linearly. For upper bound loss to be non-zero, $c_2 > \frac{\mathbbm{F}_\Theta(\bm{x}_i, \mathbbm{P})}{P_{ij}}$. Similarly, for non-zero loss, the bounds on $c_1$ are set to $\frac{\mathbbm{F}_\Theta(\bm{x}_i, \mathbbm{P})}{P_{ij}}<c_1<1$. For $c_1 = c_2 = 1$, the lower bound and upper bound are equal and hence the optimal price will be the $j^{th}$ price in the price set $\mathbbm{P}$. Therefore, $c_1$ and $c_2$ can be chosen to change the gap between the lower and upper bounds.

%% file: sections/4_priceeval.tex
\section{Pricing Model Evaluation} \label{Pricing Model Evaluation}
In the absence of optimal price values or the best hindsight strategy, it was important to the airline to define a set of offline and online metrics. Offline metrics are useful for model development, incremental learning, hyper-parameter optimization while online metrics measure business value. Establishing the exact set of offline metrics that correlates with online business metrics is an active area of research.

\subsection{Offline Metrics} \label{Offline Metrics}
In this section, we define the metrics that we use to serve as guides through hyper-parameter tuning and to ensure that nightly update of DNN weights do not overfit the data. We use the \textit{Price Decrease Recall (PDR)} and \textit{Price Decrease Precision (PDP)} scores presented by \citet{ye2018customized} due to their high correlation with the airline's business metrics. PDR measures how likely our recommended prices are lower than the current offered prices for non purchased ancillary and PDP measures the percentage for recommended prices that are lower than current offered prices for non purchased ancillary. Additionally, we use the following metrics:

\subsubsection*{Area Under the Curve (AUC)}
Due to presence of high class imbalance as discussed in Section \ref{Ancillary Purchase Prediction Model}, we used the AUC of the Receiver Operating Characteristic
(ROC) Curve as the offline metric to compare ancillary purchase prediction model performance. 
\subsubsection*{Regret Score (RS)}
In recent work \cite{ye2018customized}, regret score has been chosen as an offline evaluation criterion because of its proportional relationship to the business metric. RS is defined by equation \eqref{eq:rs}. 
\begin{equation}
\label{eq:rs}
RS = \mean_{purchases}\Big(\max\big(0, 1-\frac{P^{rec}}{P}\big)\Big)
\end{equation}

Intuitively, RS measures on an average how close our recommended price $P^{rec}$ was to the true purchase price $P$. For the example set of sessions in Table \ref{tab:example}, sessions 1, 2 and 5 have 0.20, 0.20 and 0.75 regret respectively. Because sessions 3 and 4 have recommended price higher than purchased, regret is 0. Therefore, $RS$ values for this sample of sessions is 0.095. 

\begin{table}[h]
\caption{Example prices for purchased sessions}
\begin{tabular}{@{}ccc@{}}
\toprule
Session \# & Purchase Price & Recommended Price \\ \midrule
1        & 10              & 8                 \\
2        & 15              & 12                \\
3        & 10              & 15                \\
4        & 25              & 35                \\
5        & 40              & 37                \\ \bottomrule
\end{tabular}
\label{tab:example}
\end{table}

\subsubsection*{Price Decrease F1 (PDF1)}
This score is inspired by the F1 score used to evaluate the precision and recall trade-off. PDF1 therefore measures the trade-off between PDR and PDP according to \eqref{eq:pdf1}.

\begin{equation}
PDF1 = \frac{2 \cdot PDR \cdot PDP}{PDR+PDP}
\label{eq:pdf1}
\end{equation}

\subsection{Online Metrics} \label{Online Metrics}
Online metrics represent real-world business metrics that indicate if a model is driving business value. We use two key metrics to measure the performance of a model in the real world.

\subsubsection*{Conversion Score}
One of the primary business metric is the conversion ratio, i.e., the percentage of offers that are being purchased and converted into orders. 

\begin{equation}
\text{Conversion Score} = \frac{\text{Number of purchases}}{\text{Total number of sessions}}
\end{equation}

\subsubsection*{Revenue per session}
The revenue per session ($RPS$) metric is one of the most essential business metrics to quantify actual performance. 

%% file: sections/5_Training.tex
\section{Training}\label{Training}

All of our deep neural network models (in APP-DES and DNN-CL) are trained on NVIDIA Tesla K80 GPU. We used stochastic gradient descent (SGD)\cite{bottou2010large} with a decaying learning rate to optimize the loss function. Mini-Batches and drop-out units \cite{srivastava2014dropout,glorot2010understanding} are used to regularize model training. For a discrete exhaustive search over the prices (see Section \ref{Discrete Exhaustive Search}), we use a mini-batch of the allowed price inputs to enable a single call to the GPU which minimizes data transfer and model setup cost for each price selection event. Hyperparameters like $c_1$ and $c_2$ are tuned using the bounds for non-zero loss and the upper-lower bound gap (see Section \ref{Customized Loss Function}) over the median price point in set $\mathbbm{P}$. We also use the scheduled mini-batch training approach for online model training.   

%% file: sections/6_evaluation.tex
\section{Experiments} \label{Experiments}

During the first phase of online experimentation, our airline partner's business strategy is to recommend prices equal to or less than the current human-offered price. Although this strategy reduces the search space considerably, the business motivation behind it is to reduce the overall friction in the traveler's journey by providing them an incentive to pre-purchase ancillaries online. Therefore, the aim of our online experiment is to offer discounts in an intelligent way, so that we improve the conversion rate of ancillaries without dropping the revenue per offer. This also has a potential negative impact of increasing conversion score without improving revenue per session. Therefore, as mentioned in Section \ref{Pricing Model Evaluation}, using the right set of metrics for evaluation is crucial. 

\subsection{Offline Experiments}

We perform extensive offline experiments to evaluate the performance of the models before deployment. These offline experiments consist of two parts : (i) evaluating classifiers' performance of ancillary purchase probability, and (ii) pricing effectiveness of a two-stage forecasting and optimization model versus a simultaneous, end-to-end pricing model.

\subsubsection{Classifier Performance} \label{Classifiers Performance}
It is critical to evaluate the performance of the ancillary purchase probability (APP) classifier since we are using it to estimate the demand function. Due to the high class imbalance present in the data, we used the AUC score to evaluate the performance. We started with Gaussian Naive Bayes with Clustering (GNBC) as our baseline to match the state of the art in the airline industry. We then ran the experiments with the Gaussian Naive Bayes (GNB), Random Forest (RF) and DNN classifiers, all of which performed better than the GNBC baseline. The AUC scores of $0.5716$, $0.6273$, $0.6633$ and $0.7664$ for GNBC, GNB, RF and DNN respectively, show that the DNN achieves a 33\% improvement in the AUC score compared to our baseline. This improvement is intuitive because DNNs can capture more complex relationships between the input features to predict highly imbalanced classes.

 \begin{figure}[]
\centering
\includegraphics[width=0.45\textwidth]{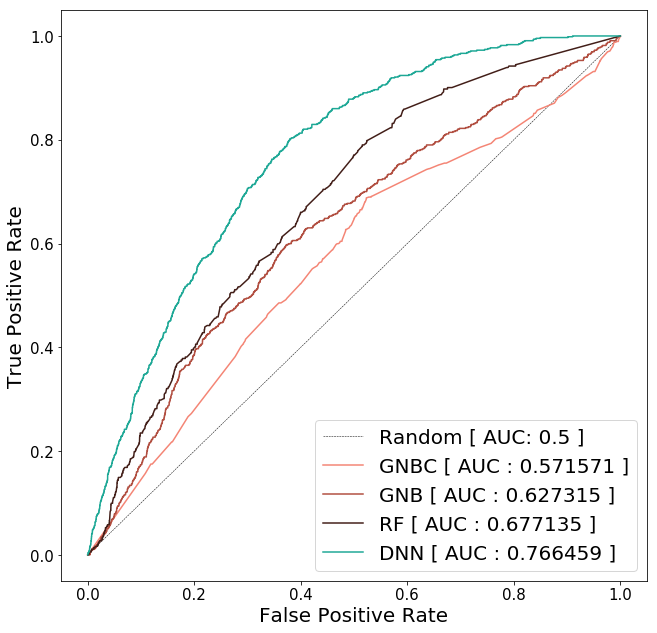}
\caption{ROC curve for offline trained model}
\label{fig:roc-auc}
\end{figure}

To evaluate the learn-ability and robustness of the classifiers, three datasets with varying amount of data are used. Datasets $A$, $B$ and $C$ have $41,000$, $50,000$ and $72,000$ sessions respectively. Results from our experiments are in Table \ref{tab:aucroc}. %Seasonal variation is introduced within the datasets by random sampling. For example, number of holidays and vacations present in dataset $C > B > A$ 
DNN shows most dominant signs of learn-ability with increasing dataset size. The best performance of these classifiers on the validation set is also presented as an ROC curve in Figure \ref{fig:roc-auc}.  
 
\begin{table}[h]
\caption{AUC score of models on datasets.}
\begin{tabular}{@{}ccccc@{}}
\toprule
Dataset & GNBC   & GNB    & RF     & DNN             \\
\midrule
A       & 0.5444 & 0.6013 & 0.6646 & \textbf{0.6755}          \\
B       & 0.5274 & 0.6186 & 0.6771 & \textbf{0.6967}          \\
C       & 0.5716 & 0.6273 & 0.6633 & \textbf{0.7664} \\
\bottomrule
\end{tabular}
\label{tab:aucroc}
\end{table}

\subsubsection{Pricing effectiveness of a sequential two-stage model versus a simultaneous end-to-end model} \label{Pricing effectiveness of a two-stage model vs. an end-end model}
Although the DNN performs well for Ancillary Probability Prediction, it was important to also measure the effectiveness of the final price recommendations from each pricing model. We used the offline metrics defined in Section \ref{Offline Metrics} to perform the comparison between the two-stage sequential forecasting and optimization models (APP-LM and APP-DES), and the simultaneous end-to-end pricing model (DNN-CL). Given the business requirement to provide discounts on the human-recommended price, we considered Regret Score (RS) and Price Decrease Recall (PDR) as more important than PDP and PDF1 \cite{ye2018customized}. Our results are summarized in Table \ref{tab:modulescore}. The APP-LM (which uses our baseline APP model and is manually tuned through a parameter search), serves as our baseline for pricing effectiveness. The inefficient performance of the APP-DES model on these metrics despite the estimation of a good APP model in the first step suggests that the price-demand relationship (see Figure \ref{fig:des}) is not estimated accurately. This shortcoming is overcome by the end-to-end model (DNN-CL), which not only overcomes the effect of this inaccuracy but also outperforms the APP-LM on all four metrics.
 
\begin{table}[h]
\caption{Comparison of scores for different models in offline experiments.}
\begin{tabular}{cccc}
\toprule
Scores & APP-LM & APP-DES & DNN-CL \\
\midrule
RS  &0.0741 & 0.3776 & 0.0726 \\
PDR &0.6366 & 0.6303 & 0.8294 \\
PDP &0.9276 & 0.9320 & 0.9230 \\
PDF1 &0.7550 & 0.7520 & 0.8737 \\
\bottomrule
\end{tabular}
\label{tab:modulescore}
\end{table}
 
%  PDR value for DNN-CL is significantly higher than APP-LM and APP-DES. Due to comparable PDP values, PDF1 score is again considerably higher for DNN-CL. 
Hence, we conclude that the DNN-CL model not only minimizes the regret for not pricing high for purchased ancillaries, but also maximizes the likelihood of the recommended prices being low when ancillaries are not purchased.

\subsection{Online Experiments} \label{Online Experiments}

Our APP-LM model has been deployed in production on our partner airline's internet booking engine for model validation. The APP-DES and the DNN-CL models are currently under deployment, following their successful performance according to the offline metrics.

According to the airline's business strategy, we introduced a random discount model in addition to the APP-LM model. This random discount model is allowed to recommend discounted prices based on Gaussian noise. There are two reasons for deploying a random price recommender. First, it establishes a baseline for conversion score improvements from discounted ancillaries. Second, it enables us to explore various prices and calibrate price sensitivity. The deployed models are compared with both human-curated static prices and prices from the random discount model. All three were deployed concurrently in an A/B testing setting for a period of 120 days. The results of this comparison for the most recent 30 days are shown in Table \ref{tab:RPO}.

\begin{table}[h]
  \caption{Conversion percentage and revenue generated by our model (APP-LM) compared to human-curated and random prices}
  \label{tab:RPO}
  \begin{tabular}{ccc}
    \toprule
    Pricing System         & Avg. Revenue per Offer  & Conversion Score\\
    \midrule
    HUMAN           & 1.00    & 10.18\%    \\
    RANDOM  & 0.77 & 12.37\%    \\
    APP-LM          & 1.10 & 13.92\%    \\
  \bottomrule
\end{tabular}
\label{tab:onlinecomparison}
\end{table}

\begin{figure}[h]
\centering
\includegraphics[width=0.45\textwidth]{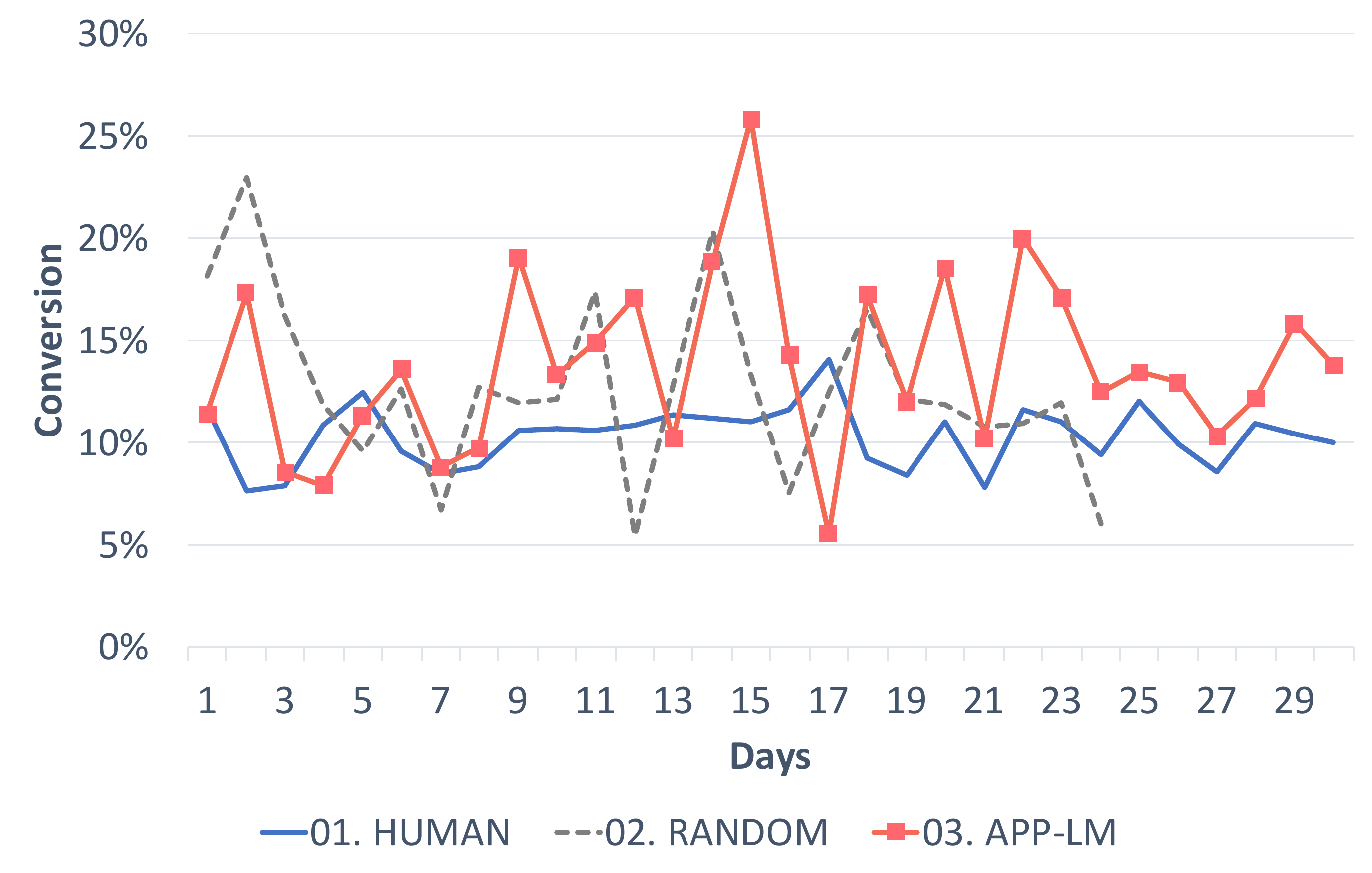}
\caption{Conversion Score by Each Pricing System Over Time}
\label{fig:conversionperuser}
\end{figure}

Figure \ref{fig:conversionperuser} and Table \ref{tab:onlinecomparison} indicate that the random discount model is able to produce higher conversion rates than the human-curated pricing system. This not only demonstrates the existence of price sensitivity among customers, but also allows us to measure it. Additionally, the random discount model is unable to produce higher revenue per offer because it conflates the sub-markets' demand and makes them indistinguishable, thus losing information.

\begin{figure}[h]
\centering
\includegraphics[width=0.45\textwidth]{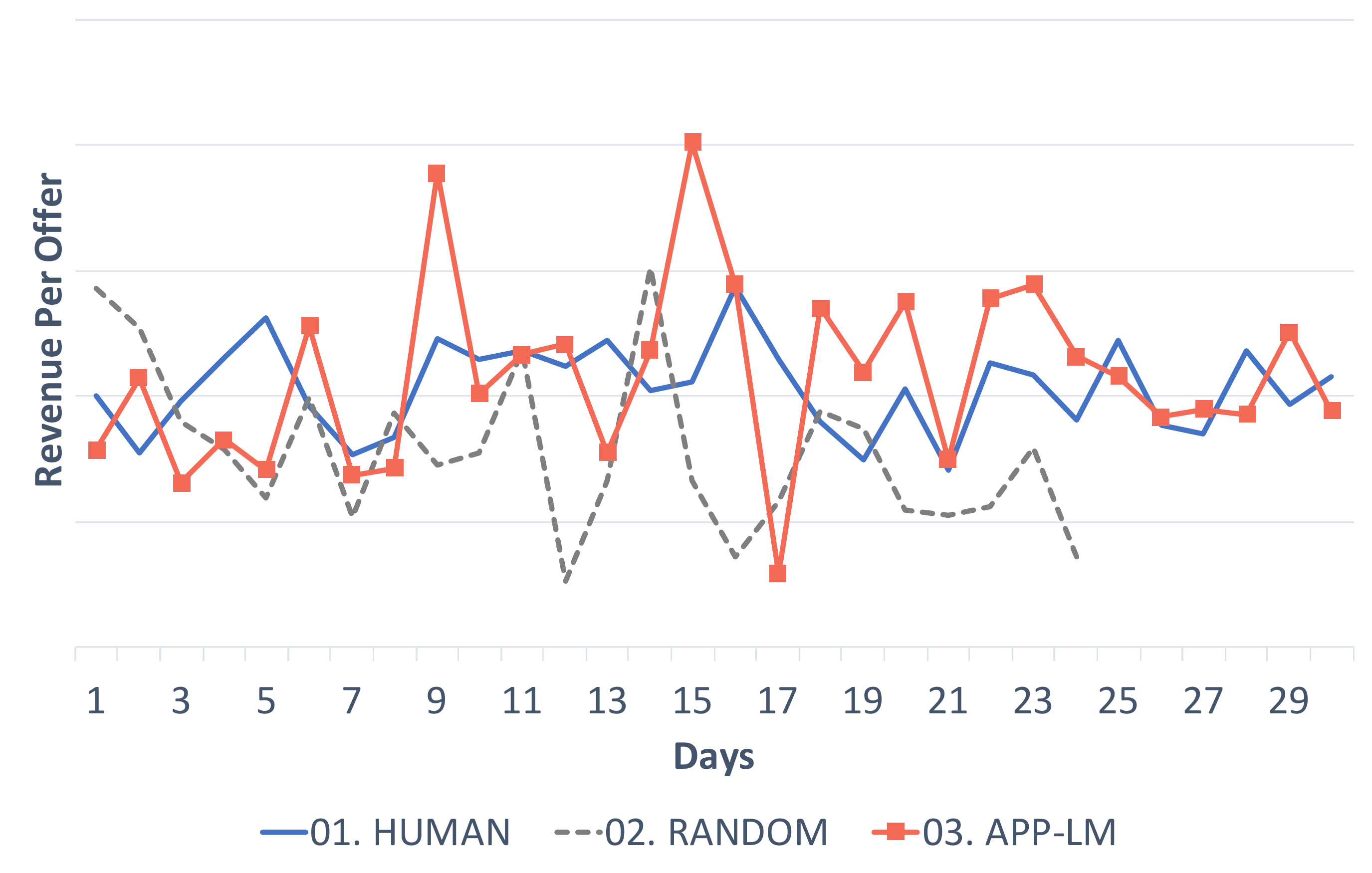}
\caption{Revenue Per Offer by Each Pricing System Over Time}
\label{fig:revenueperuser}
\end{figure}

Because customers' price sensitivity is observed through the random discount model, a slight increase in the conversion score was expected in our deployed model APP-LM. The results in Table \ref{tab:onlinecomparison}, Figure \ref{fig:revenueperuser} and Figure \ref{fig:conversionperuser}\footnote{The exact dates and revenue figures cannot be included due to proprietary, privacy and sensitivity restrictions} confirm this expectation. We observe a $36\%$ increase in conversion rate, which is a $15\%$ increase compared to the random discount model. More importantly, our model produces $10\%$ more revenue than the human-curated pricing system. This implies that our model recommends lower prices to targeted customers such that the revenue per offer from our model can still outperform (or at least be comparable to) the human-curated pricing system. For revenue per offer and conversion score, we see that our model can indeed capture the market trend in a timely fashion. Furthermore, the clear trend of both higher revenue and higher conversion score with respect to the human-curated system indicate the accuracy of target discount with the customer's context. 
\pagebreak

% on a daily basis.

%% file: sections/7_discussion.tex
\section{Future Work}
Historically, price sensitivity to ancillaries has not been captured due to static pricing. However, it is critical to capture customers' price sensitivity to price the product correctly and maximize revenues. Currently, all of our proposed models - APP-LM, APP-DES and DNN-CL - are in early exploratory stage to capture and train on the ground truth responses of customers. While APP-LM has been deployed online, APP-DES and DNN-CL are currently being deployed. Once model validation is performed online, the airline's booking system will switch to the most robust model. For the APP-DES and DNN-CL models, we are specifically interested in further examining the correlation between offline model performance to online business performance, because they outperform APP-LM in offline experimentation. Further, our deployment system will be transitioned from a scheduled mini-batch training (see Section \ref{Training}), to an event-wise online training. This transition will enable the model to accurately learn temporal dependencies. We also plan to alter the current business strategy (see Section \ref{Experiments}) to allow our model to recommend prices higher then current limit, and observe customer responses. Finally, we plan to study the effect of heterogeneous ancillary types being dynamically priced by our deployed models, and the best predicted subset of ancillaries being offered to the customer. Given that various ancillaries compete for wallet-share and shelf-space, it will help expand our understanding of whether such pricing models compete, or collaborate, with each other.

%% file: sections/8_conclusion.tex
\section{Conclusion}
In this paper, we present a first step in the direction of efficient inference systems from booking data in the airline industry, compared to past works that focus on strategic impacts. We successfully demonstrate that ancillaries can be dynamically priced without using any user specific information that violates customer privacy. We compared three different dynamic pricing models (APP-LM, APP-DES and DNN-CL) and their associated frameworks. Our results show that the accuracy of estimating the demand and fine-tuning its sensitivity to price, greatly influences the optimality of the recommended price. Our offline experiments indicate that DNN-CL can perform significantly better than APP-LM, APP-DES, and currently deployed approaches, to maximize revenue. In online experiments, our APP-LM model outperforms human-curated pricing systems currently in use. By using reliable evaluation metrics that correlate well with business impact, we hope to observe further improvement in online metrics through our APP-DES and DNN-CL models that are currently under deployment. Our work demonstrates the promise of improved business value through highly accurately, continuously updated models for customer demand for ancillaries, and their sensitivity to prices.